\pdfoutput=1

\documentclass[11pt]{article}

\usepackage{graphicx}
\usepackage{colortbl}
\usepackage{xcolor,color}
\usepackage{multirow}
\usepackage{acl}

\usepackage{times}
\usepackage{latexsym}
\usepackage{tipa}

\usepackage[T1]{fontenc}

\usepackage[utf8]{inputenc}

\usepackage{microtype}

\newcommand*\rot{\rotatebox{90}}

\title{Approaching Reflex Predictions as a Classification Problem Using Extended Phonological Alignments}

\author{Tiago Tresoldi \\
  Department of Linguistics and Philology \\
  University of Uppsala \\
  Uppsala \\
  \texttt{tiago.tresoldi@lingfil.uu.se} 
  }

\begin{document}
\maketitle

\begin{abstract}
This work describes an implementation of the ``extended alignment'' (or ``multitiers'') approach for cognate reflex prediction, submitted to ``Prediction of Cognate Reflexes'' shared task. Similarly to \citealt{List2022d}, the  technique involves an automatic extension of sequence alignments with multilayered vectors that encode informational tiers on both site-specific traits, such as sound classes and distinctive features, as well as contextual and suprasegmental ones, conveyed by cross-site referrals and replication. The method allows to generalize the problem of cognate reflex prediction as a classification problem, with models trained using a parallel corpus of cognate sets. A model using random forests is trained and evaluated on the shared task for reflex prediction, and the experimental results are presented and discussed along with some differences to other implementations.
\end{abstract}

\section{Introduction}

The Special Interest Group of Linguistic Typology (SIGTYP) organized in 2022 the ``Prediction of Cognate Reflexes'' shared task \cite{List2022g}, providing the community with  cognate-coded wordlists from which varying amounts of cognate sets were withheld. Participants were asked to submit models capable of predicting the missing words and morphemes from the non-withheld members. This work describes the submission named \texttt{ceot-extalign-rf}.

Reflex prediction is particularly interesting to computational historical linguistics, since most approaches to reconstruction are rooted in comparative methods operating on cognate sets \cite{Jaeger2019,greenhill2020bayesian,list2018sequence}, i.e., sets of words that are assumed to derive from a common proto-word. We can therefore define a cognate as a member of a set of words (a ``cognate set'') that share an etymological origin via vertical descent. Such a definition highlights how the defining property of a cognate set is the regular sound correspondences between the words involved, even when they are not judged as ``similarly sounding'' by human evaluators. Homologies don't imply homogenies, as processes like word borrowing and chance resemblance can lead to similarities, while true cognates can be very dissimilar, as in the often cited example of English ``two'' and Armenian ``erku'', both from a reconstructed form *dwóh$_1$ \cite{kroonen2013etymological}.

The task can be seen as a form of zero-shot learning \citep{xian2018zero}, where a model must learn to predict the ``reflexes'' of a potentially unknown ancestral word form, with no examples of the relevant cognate set provided during the training phase. When considering the landscape of machine learning methods available and the approaches so far proposed \citep{Dinu2014,Bodt2021,Meloni2021,Beinborn2013,Dekker2021,Fourrier2021,List2022d}, including other submissions to this challenge \cite{Jaeger2022,Celano2022,Kirov2022}, it is possible to identify two main strategies for the task. The first one treats the problem as one of classification, potentially refining sequence results with probabilities from a character model, while the second employs sequence transformation methods, especially those akin to  \textit{seq2seq} approaches \cite{sutskever2014sequence}, making the task one analogous to that of ``translation''. This submission is of the first kind, with results provided by random forests \cite{kam1995random,hastie2009elements}, but attention should focus on the proposed method of data transformation.

The method is based on ``multitiered'' proposals, originally based on an idea by J.-M. List, which have been described and implemented in other works \cite{Tresoldi2018,List2019PBLOG4,Bodt2021,ChaconList2016}, including for the baseline for the shared task at hand \cite{List2022e}. It is an approach for modeling historical linguistic relationships developed to solve problems that are not addressed by pure-correspondence approaches, such as those involving cross-site generalization and the need to capture contextual and suprasegmental information. By considering how discrete representations of phonological sequences can be an unsuitable provision, as the latter are abstract representations of both discrete and continuous multidimensional phonological domains, the technique extends base alignments with multilayered vectors, encoding additional and derived features. Here, such an informational extension is knowingly and intentionally similar to the analytical frameworks of Firthian \cite{mitchell1975} and autosegmental phonology \cite{goldsmith1990autosegmental}, where sequence representation is given by more than a single string of segments.

This work first summarizes the extended alignment technique, with a focus on how it can be used to predict cognate reflexes. Then, it describes the experimental setting used for the submission, and how the extended alignment method can apply to the task at hand. It concludes with a discussion on the results and future work.

\section{Materials and Methods}

\subsection{Materials}

Data for the experiment comprised 20 standardized datasets derived from the Lexibank project \cite{List2022e}, each encompassing a single linguistic family and providing good geographic and typological variation. The datasets were split into five partitions each, with different ratios of words kept for training and evaluation, and were provided by the task's organizers. They provided a more comprehensive description of the data and the way they prepared it in \citet{List2022g}.

The submission also extensively uses the phonological information provided by the \textit{mipa} and \textit{tresoldi} models of the \texttt{maniphono} library \citep{Tresoldi2021maniphono}, which were incorporated into the code in order to simplify installation requirements. The information provided by these models is adapted from the data of CLTS \citep{Anderson2018,CLTS}, with additional mappings and structures partly described in \citet{tresoldi_calc}.

\subsection{Methods}

\subsubsection{Multitiered Extension}

It is important to distinguish between the method for representing phonological data and the actual model for reflex prediction. The first, building upon the theoretical discussions also shown in \citet{List2022e}, is the most promising element because of its innovative treatment of alignment sites: instead of just being  linear components of a sequence under an alignment set, they are treated as independent records in a database. Such a database is turned into a two-dimensional matrix, from which observed and predicted features are extracted and used by common classification models. The method thus facilitates the usage of established algorithms for machine learning, allowing to reframe tasks from historical linguistics as more common tasks of classification, regression, or transformation by using conventional methods and well-researched implementations.

Let's consider the example given in the challenge's call, reproduced in Table \ref{tab:cogs}, with three cognate sets, identified by comparable concepts, involving German, English, and Dutch.

\begin{table}[htb]
    \centering
    \begin{tabular}{|l|l|l|l|}\hline
Cognate Set & German & English & Dutch \\\hline
ASH & \textipa{a S E} & \textipa{\ae S} & \textipa{A s} \\
BITE & \textipa{b ai s @ n}  & \textipa{b ai t} & \textipa{b Ei t @} \\
BELLY & \textipa{b au x} & ? & \textipa{b {\oe}i k}  \\\hline
    \end{tabular}
    \caption{Exemplary cognate reflexes in German, English, and Dutch. Adapted from \citet{List2022g}.}
    \label{tab:cogs}
\end{table}

The first step in producing an alignment enriched with new tiers is to perform multiple sequence alignment \citep{List2012c}, which in this case yields three independent alignment sets, as shown in Tables \ref{tab:alm1}, \ref{tab:alm2}, and \ref{tab:alm3}. This is a step common to most classification methods for reflex prediction.

\begin{table}[htb]
    \centering
    \begin{tabular}{|l|c|c|c|}\hline
    Language & \#1 & \#2 & \#3 \\\hline
German  & \textipa{a}   & \textipa{S} & \textipa{E}  \\
English & \textipa{\ae} & \textipa{S} & -  \\
Dutch   & \textipa{A}   & \textipa{s} & -  \\\hline
    \end{tabular}
    \caption{Alignment for cognate set ASH for German, English, and Dutch reflexes.}
    \label{tab:alm1}
\end{table}

\begin{table}[htb]
    \centering
    \begin{tabular}{|l|c|c|c|c|c|}\hline
    Language & \#1 & \#2 & \#3 & \#4 & \#5 \\\hline
German  & \textipa{b} & \textipa{ai} & \textipa{s} & \textipa{@} & \textipa{n} \\
English & \textipa{b} & \textipa{ai} & \textipa{t} & -           & -\\
Dutch   & \textipa{b} & \textipa{Ei} & \textipa{t} & \textipa{@} & -\\\hline
    \end{tabular}
    \caption{Alignment for cognate set BITE for German, English, and Dutch reflexes.}
    \label{tab:alm2}
\end{table}

\begin{table}[htb]
    \centering
    \begin{tabular}{|l|c|c|c|}\hline
    Language & \#1 & \#2 & \#3 \\\hline
German  & \textipa{b} & \textipa{au} & \textipa{x} \\
English & ? & ? & ? \\
Dutch   & \textipa{b} & \textipa{{\oe}i} & \textipa{k} \\\hline
    \end{tabular}
    \caption{Alignment for cognate set BELLY for German, English, and Dutch reflexes.}
    \label{tab:alm3}
\end{table}

Even without extensions, by considering each alignment site an independent observation it is possible to combine multiple sets into a single data frame. The operation ``transposes'' the  alignments, joining them into a single frame as shown in Table \ref{tab:tiers1}, following the common steps of this framework \cite{List2019a,Tresoldi2018,Bodt2021,List2022e}.

\begin{table}[htb]
    \centering
    \begin{tabular}{|l|c|c|c|c|}\hline
    \rot{ID} & \rot{Source} & \rot{German segment} & \rot{English segment} & \rot{Dutch segment} \\\hline
1  & ASH.1   & \textipa{a}  & \textipa{\ae}  & \textipa{A}  \\
2  & ASH.2   & \textipa{S}  & \textipa{S}    & \textipa{s}  \\
3  & ASH.3   & \textipa{E}  & -              & -            \\
4  & BITE.1  & \textipa{b}  & \textipa{b}    & \textipa{b}  \\
5  & BITE.2  & \textipa{ai} & \textipa{ai}   & \textipa{Ei} \\
6  & BITE.3  & \textipa{s}  & \textipa{t}    & \textipa{t}  \\
7  & BITE.4  & \textipa{@}  & -              & \textipa{@}  \\
8  & BITE.5  & \textipa{n}  & -              & -            \\
9  & BELLY.1 & \textipa{b}  & ?              & \textipa{b}  \\
10 & BELLY.2 & \textipa{au} & ?              & \textipa{{\oe}i}  \\
11 & BELLY.3 & \textipa{x}  & ?              & \textipa{k}       \\\hline
    \end{tabular}
    \caption{Extended data frame, without any contextual extension, from the aligned German, English, and Dutch reflexes for the cognate sets ASH, BITE, and BELLY. Note that ``Source'' is only provided here for ease of exposition, and is not part of actual implementation.}
    \label{tab:tiers1}
\end{table}

Such an organization of the data is already appropriate for training statistical methods, as each observation is independent, making it possible to identify correspondences and fill gaps by imputing values. For instance, the partial match between the sites of index 4 and 9 in Table \ref{tab:tiers1} strongly suggests that we should impute the missing information for the English segment as \textipa{/b/}, or that site 11 should be a dorsal consonant. Despite disappointing performance in most cases given by the lack of information for a proper zero-shot classification, reflex prediction would already be possible due to the comparatively high number of informative features extracted from a small set of alignments, especially if the prediction refers to phonological traits instead of atomic segments. For example, if given a hypothetical partial cognate set with a German reflex \textipa{/Saus/} and an English \textipa{/Sout/}, most statistical methods could already classify the first site in the alignment as analogous to site 2 (given the \textipa{/S/} to \textipa{/S/} correspondence between German and English), the second one to site 10 (even though English \textipa{/ou/} is not attested in this example), and the third one to site 6 (given the \textipa{/s/} to \textipa{/t/} correspondence), yielding a hypothetical Dutch form \textipa{/s{\oe}it/}.

The data frame can be extended in two ways. First, it can be enriched with information specific to each alignment site, allowing machine learning methods to generalize from observed instances (for example, learning that a correspondence applies not just to one sound, but to one or more sets of sounds) and to restrict the effect of correspondences to certain word or syllable positions\footnote{Note that word position of the alignment site is explicitly tested by \citet{List2022d}, which report negligible improvements in performance.}. In Table \ref{tab:tiers2}, such features are added by extending sites with tiers for the ``sound class'' \cite{Dolgopolsky1986} under the SCA model \cite{List2012c} and the alignment position for each segment. In an actual implementation, more site-specific information would likely be added, such as tiers derived from distinctive features (both from commonly used models, like those derived from \citealt{chomsky1968sound}, and from binary models designed for machine learning, like those provided by \texttt{maniphono}) and indexes related to the position in the word and in the syllable when counting either left-to-right (``index'') and right-to-left (``rindex'').

\begin{table}[htb]
    \centering
    \begin{tabular}{|l|c|c|c|c|c|c|c|c|}\hline
    \rot{ID} & \rot{Index} & \rot{German segment} & \rot{German SC} & \rot{English segment} & \rot{English SC} & \rot{Dutch segment} & \rot{Dutch SC} \\\hline
1  & 1 & \textipa{a}  & A & \textipa{\ae}  & E & \textipa{A}      & A \\
2  & 2 & \textipa{S}  & S & \textipa{S}    & S & \textipa{s}      & S \\
3  & 3 & \textipa{E}  & E & -              & - & -                & - \\
4  & 1 & \textipa{b}  & P & \textipa{b}    & P & \textipa{b}      & P \\
5  & 2 & \textipa{ai} & A & \textipa{ai}   & A & \textipa{Ei}     & E \\
6  & 3 & \textipa{s}  & S & \textipa{t}    & T & \textipa{t}      & T \\
7  & 4 & \textipa{@}  & E & -              & - & \textipa{@}      & E \\
8  & 5 & \textipa{n}  & N & -              & - & -                & - \\
9  & 1 & \textipa{b}  & P & ?              & ? & \textipa{b}      & P \\
10 & 2 & \textipa{au} & A & ?              & ? & \textipa{{\oe}i} & U \\
11 & 3 & \textipa{x}  & G & ?              & ? & \textipa{k}      & K \\\hline
    \end{tabular}
    \caption{Multitiered representation extended with information on the alignment index and the corresponding SCA sound class, from the aligned German, English, and Dutch reflexes for sets ASH, BITE, and BELLY.}
    \label{tab:tiers2}
\end{table}

The second type of extensions addresses the fact that making each alignment site independent loses contextual information. In Table \ref{tab:tiers3},  two contextual tiers are added for each segment tier, one specifying the previous segment (the segment one position to the left, thus L1) and one carrying information on the following sound class (the SCA one position to the right, thus R1). There is no limit on the amount of contextual information with which each alignment site can be enriched, and, in fact, when using a complete system of phonological features it is possible to encode complex phonological information such as ``the preceding syllable has a nasal consonant'' or ``the word ends with a front vowel''.

\begin{table*}[htb]
    \centering
    \begin{tabular}{|l|c|c|c|c|c|c|c|c|c|c|c|c|c|c|}\hline
    \rot{ID} & \rot{Index} &
    \rot{German segment} & \rot{German SC} & \rot{German segment L1} & \rot{German SC R1} &
    \rot{English segment} & \rot{English SC} & \rot{English segment L1} & \rot{English SC R1} &
    \rot{Dutch segment} & \rot{Dutch SC} & \rot{Dutch segment L1} & \rot{Dutch SC R1} 
    \\\hline
1  & 1 & \textipa{a}  & A & \O           & S  & \textipa{\ae}  & E & \O            & S  & \textipa{A}      & A & \O               & S  \\
2  & 2 & \textipa{S}  & S & \textipa{a}  & E  & \textipa{S}    & S & \textipa{\ae} & -  & \textipa{s}      & S & \textipa{A}      & -  \\
3  & 3 & \textipa{E}  & E & \textipa{S}  & \O & -              & - & \textipa{S}   & \O & -                & - & \textipa{s}      & \O \\
4  & 1 & \textipa{b}  & P & \O           & A  & \textipa{b}    & P & \O            & A  & \textipa{b}      & P & \O               & E  \\
5  & 2 & \textipa{ai} & A & \textipa{b}  & S  & \textipa{ai}   & A & \textipa{b}   & T  & \textipa{Ei}     & E & \textipa{b}      & T  \\
6  & 3 & \textipa{s}  & S & \textipa{ai} & E  & \textipa{t}    & T & \textipa{ai}  & -  & \textipa{t}      & T & \textipa{Ei}     & E  \\
7  & 4 & \textipa{@}  & E & \textipa{s}  & N  & -              & - & \textipa{t}   & -  & \textipa{@}      & E & \textipa{t}      & -  \\
8  & 5 & \textipa{n}  & N & \textipa{@}  & \O & -              & - & -             & \O & -                & - & \textipa{@}      & \O \\
9  & 1 & \textipa{b}  & P & \O           & A  & ?              & ? & \O            & ?  & \textipa{b}      & P & \O               & U  \\
10 & 2 & \textipa{au} & A & \textipa{b}  & G  & ?              & ? & ?             & ?  & \textipa{{\oe}i} & U & \textipa{b}      & K  \\
11 & 3 & \textipa{x}  & G & \textipa{au} & \O & ?              & ? & ?             & \O & \textipa{k}      & K & \textipa{{\oe}i} & \O \\
\hline
    \end{tabular}
    \caption{Data frame of alignment sites extended with information on the alignment index, the corresponding SCA sound class, the preceding segment, and the following SCA sound class, from the aligned German, English, and Dutch reflexes for the cognate sets ASH, BITE, and BELLY. Missing information, such as the segment to the left of the first site in an alignment, is marked with \O.}
    \label{tab:tiers3}
\end{table*}

Depending on the size of the context window, the number, and the type of tiers used for extending the alignment, the shape of the data frame can increase to hold hundreds or thousand of features. While the human inspection of such data will rapidly become impractical, this property should not be considered an issue because, at this stage, the representation is intended for machine consumption. Methods for extracting information for human consumption should rely on these enlarged data frames, without reducing their size beforehand. Nonetheless, before carrying any kind of statistical analysis, it is recommended to perform common tasks of data preparation, such as dimensionality reduction and scaling of the features. Data standardization and normalization can ensure that features extracted from different tiers are placed on a common scale, and transformation processes such as Principal Component Analysis (PCA) \cite{tipping1999probabilistic} can be improve training times and performance.

The implementation presented in this work differs from the previous ones due to the greater attention to the principles of autosegmental phonology and, as already mentioned, for allowing the use of the propagation strategy of contextual information also for suprasegmental features, such as stress and tone, to all alignment sites where it applies. With tones, for example, instead of marking the tone as a segment-like token at the end of the syllable or as a property of the nucleus alone, it is possible to expand the alignment with one or more tiers regarding the relevant tonality, which will apply to all the segments that make up the relevant syllable. Such information is not restricted to the tone itself, but can be decomposed into properties like ``tone contour'' or ``starting pitch''. With complex correspondences involving suprasegmental features, machine learning methods will not need to ``look ahead'' for a tone token, as one or more columns will carry the relevant information in the data frame record itself. In addition, the representation structure allows more easily composing results from different information tiers: instead of establishing models that only predict segments, as atomic units, the implementation allows to predict, contemporaneously or individually, two or more tiers. whose information can be combined for the final results. For example, especially with a binary model, it is possible to predict the manner and place of articulation of a segment independently, aggregating the results into a phoneme or sound class.

\subsubsection{Cognate Prediction}
\label{subsec:cogpred}

Cognate prediction is performed by training classifiers on the data frames prepared by the code for extending alignments. When paying attention to issues such as scaling of features, missing data, and the encoding of multistate categorical features (usually with one- or multi-hot binary encoders), any machine learning method can be used.

The task will involve two subtasks: the first for training all the classifiers that are needed, and the second for generating output in the expected format once the classifiers have been prepared.

The steps for the first subtask are:

\begin{enumerate}
    \item \textbf{Align raw data}. This step can be performed manually or with tools for linguistic alignments, such as LingPy \cite{LingPy}.
    \item \textbf{Prepare extended data frames}. From the alignment sets, a single data frame is generated with all the requested additional tiers, both for in-site and contextual information. 
    \item \textbf{Prepare the training data for each language}. The training data comprises a data frame for input variables \textit{X}, including all features save for those related to the language being predicted, and a vector for the output variable \textit{y}, from the appropriate ``segment'' column. All other features related to the language under study are discarded.
    \item \textbf{Train and save classifiers}. This will result in a collection with one classifier for each language in the dataset.
\end{enumerate}

Once the classifiers are ready, it is possible to perform  reflex prediction with the following steps:

\begin{enumerate}
    \item \textbf{Align raw data}. As above.
    \item \textbf{Prepare extended data frames}. As above; it is highly recommended, and depending on the machine learning method it is necessary, that the set of tiers added to the alignment is equal to or a subset of the one used in training.
    \item \textbf{Prepare the \textit{X} data frame and generate a \textit{y} prediction}. For most classification methods, \textit{y} will yield a probability for different segments.
    \item \textbf{Prepare the output}. Build the sequences of predicted reflexes and organize them in the expected data structure for evaluation.
\end{enumerate}

\section{Implementation and Results}

For the shared task, data and classifiers were prepared according to the workflows described in subsection \ref{subsec:cogpred}.

Due to design decisions aiming at testing the method more than achieving the best performing model, only few additional tiers were used for extending the base alignments. These were the segments and SCA sound classes pertinent to each language, with a left and right order of 1 and 2 (i.e., L1, L2, R1, R2), thus increasing four times the number of phonological tiers. Indexing tiers related to the position in the alignment, counting both left-to-right and right-to-left, were also added. No pruning or preemptive dimensionality reduction was performed. Random forests \citep{breiman2001random,kam1995random,hastie2009elements} were trained using the default implementation in \textit{scikit-learn} version 1.1.0 \citep{scikit-learn}. Other classification methods were explored using samples of the datasets, yielding good performance improvements  in the cases of XGBoost \citep{Chen:2016:XST:2939672.2939785},
LightGBM \citep{NIPS2017_6449f44a}, and multi-layer perceptrons \cite{hinton1990connectionist}, particularly when performing hyperparameter optimization \cite{10.1145/3292500.3330701}. The decision to only submit the results using the random forest method was based on the time and computational constraints to perform a full training, as well as in the goal of establishing a baseline for this implementation of the method.

Comparing with both the baseline and other submitted methods, the final results suggest much room for improvement. Performance was in general between the Baseline and Baseline-SVM method, mostly due to gains in prediction offered by Support Vector Machines \cite{koutroumbas2008pattern}, for the partitions with most information, degrading rapidly when the amount of withheld data was increased, as illustrated by the global results reported in \citet{List2022g}.

The implementation performed particularly poorly with the datasets \texttt{bantudvd} \cite{Greenhill2015a} (as illustrated by the 0.7053 B-Cubes for the 10\% partition, versus 0.7835 of the Baseline submission),
\texttt{felekesemitic} \cite{Feleke2021} (0.6661 B-Cubes for the 10\% partition, versus 0.6925 of the baseline). It performed better with datasets
\texttt{beidazihui} \cite{Zihui} (as illustrated by the 0.8356 B-Cubes for the 10\% partition, versus 0.7279 of the baseline),
\texttt{bodtkhobwa} \cite{Bodt2021} (0.7993 B-Cubes for the 10\% partition, versus 0.7566 of the baseline),
\texttt{bremerberta} \cite{Bremer2016} (0.7915 B-Cubes for the 10\% partition, versus 0.7187 of the baseline),
\texttt{deepadungpalaung} \cite{Deepadung2015} (0.8143 B-Cubes for the 10\% partition, versus 0.7597 of the baseline),
\texttt{wangbai} \cite{Wang2004} (0.8326 B-Cubes for the 10\% partition, versus 0.8048 of the baseline),
\texttt{hattorijaponic} \cite{Hattori1973} (0.8127 B-Cubes for the 10\% partition, versus 0.7889 of the baseline),
\texttt{listsamplesize} \cite{List2014c} (0.5325 B-Cubes for the 10\% partition, versus 0.4048 of the baseline). Full results are available along with the submission, with performance for other datasets comparable to the baseline.

\section{Discussion}

Similarly to the implementation in \citealt{List2022d}, the method of extending alignments implemented here can be conceived as a learned data augmentation strategy, since it is not based on statistical properties of each dataset (that is, each collection of alignment sets), but on the contribution from specialized linguistic knowledge. Such a strategy does not exclude the use of purely statistical methods, allowing them to more easily find the correspondences between sets, especially when they convey phonological traits that are sparse and non sequential. Once mature, we believe that the strategy will be beneficial for most machine learning methods, as it should allow for increased generalization due to increased complexity of the data sets, incorporating informational tiers beyond phonological or sequence properties. 

Extended alignments have proven their usefulness for tasks like supervised phonological reconstruction and cognate reflex prediction. There are, however, many other problems to which they could also be applied, either by replacing or by complementing existing methods. The probability of correspondence between certain phonemes, for example, can be used for fine-grained decisions in cases of doubts in the domain of sequence alignment, using information which is ``local'' to the languages under study along with default correspondence score matrices that are generally computed for global usage. Likewise, the method can credibly be used in the detection of loanwords \cite{miller2020}, especially in cases of large exchange of words between two languages and of ``learned loanwords''. Finally, like the other models proposed for this challenge, the method can be used to manage linguistic data, identifying forms that are not predicted by a model trained on the data itself, which may be due to either  particularly rich individual histories or different amounts data noise.

\section*{Acknowledgements}

The author is supported by the Cultural Evolution of Texts project, with funding from the Riksbankens Jubileumsfond (grant agreement ID: MXM19-1087:1). Theoretical work was partly developed when the author received funding from the European Research Council (ERC) under the European Union’s Horizon 2020 research and innovation programme (grant agreement No. \#715618, ``Computer-Assisted Language Comparison'').

\section*{Supplementary Material}

Code and data for reproducing the analyses is deposited at \url{https://doi.org/10.5281/zenodo.6563490}.


\bibliography{anthology}
\bibliographystyle{acl_natbib}



\end{document}